\documentclass[journal]{IEEEtran}
\usepackage{cite}
\ifCLASSINFOpdf
  \usepackage[pdftex]{graphicx}
\else
\fi
\usepackage[cmex10]{amsmath}
\usepackage{array}
\usepackage{fixltx2e}
\usepackage{stfloats}
\fnbelowfloat

\usepackage{pgf}
\usepackage{tikz}
\usetikzlibrary{arrows,positioning}

\usepackage{url}
\usepackage{color}

\usepackage{color}

\begin{document}
\title{An evaluation framework for event detection using a morphological model of acoustic scenes}

\author{
		Mathieu~Lagrange,~\IEEEmembership{IEEE memberships},
		Gr\'egoire~Lafay,
        Mathias~Rossignol,~\IEEEmembership{IEEE memberships},    
        Emmanouil Benetos,~\IEEEmembership{Member, IEEE}, % <-this % stops a space
        Axel~Roebel,~\IEEEmembership{IEEE memberships},
     %   Mark Plumbley,~\IEEEmembership{IEEE memberships} % <-this % stops a space
\thanks{ANR Houle under reference ANR-11-JS03-005-01. EB is supported by a City University London Research Fellowship.}% <-this % stops a space
\thanks{}% <-this % stops a space
\thanks{}
}


\markboth{DRAFT - JOURNAL OF XXXX, VOL. XX, NO. XX, XX 20XX}%
{Bare Demo of IEEEtran.cls for Journals}



\maketitle

\begin{abstract}

This paper introduces a model of environmental acoustic scenes which adopts a morphological approach by abstracting temporal structures of acoustic scenes. To demonstrate its potential, this model is employed to evaluate the performance of a large set of acoustic events detection systems. This model allows us to explicitly control key morphological aspects of the acoustic scene and isolate their impact on the performance of the system under evaluation. Thus, more information can be gained on the behavior of evaluated systems, providing guidance for further improvements. The proposed model is validated using submitted systems from the IEEE DCASE Challenge; results indicate that the proposed scheme is able to successfully build datasets useful for evaluating some aspects the performance of event detection systems, more particularly their robustness to new listening conditions and the increasing level of background sounds.

\end{abstract}

% Note that keywords are not normally used for peerreview paper.

% \fi
\IEEEpeerreviewmaketitle

\section{Introduction}

\IEEEPARstart{O}{ver} the past decades, the amount of audio data in our sonic environment have considerably grown. Recent research fields such as eco-acoustics \cite{ECOACOUSTICS2014, krause} start to massively record environmental sounds around the world in order to measure potential animal biodiversity modification over large temporal scales due to human activity or climate change \cite{NessSST13, stowell13a, stowell13b}. Other research fields focus on human activities for context inference and surveillance \cite{heittola13,1540194,Heittola2013,park14}.

If research on Automatic Speech Recognition (ASR) \cite{Rabiner93} and Music Information Retrieval (MIR) \cite{Muller07} are now well established, research addressing automatic analysis of complex environmental acoustic scenes remains relatively young. In particular, those open research avenues suggest a large range of experimentation in order to 1)  gain knowledge about the important characteristics of those acoustic scenes and how they can be modeled, 2) propose new algorithmic approaches to contribute to the above cited applications area: eco-acoustics and urban sensing. Being relatively new research fields, only few data sets are available, though this number may grow as the interest of scientific and engineering communities for such tasks increases, see \cite{Giannoulis:2013a} and \cite{salamon14} for research effort in human environments related tasks.

This paper  focuses on building an evaluation framework for the task of detecting events of interest in acoustic scenes using simulated data. It builds upon the IEEE AASP Challenge on Detection and Classification of Acoustic Scenes and Events (DCASE), which was organised by the Centre for Digital Music of Queen Mary University of London and by the Institute for Research and Coordination in Acoustics/Music (IRCAM), under the auspices of the Audio and Acoustic Signal Processing (AASP) technical committee of the IEEE Signal Processing Society in 2013 \cite{giannoulis2013detection}. The DCASE Challenge is the second challenge dedicated to this task after the CLEAR Challenge \cite{clear}.

During the formal definition of the event detection task of this challenge, besides important questions about evaluation metrics, an interest rose about the potential benefit of considering simulated data to enlarge the scope of evaluation of the submitted systems. Varying the power level of the background, the density of the events, their intra class diversity, all seemed important aspects to would be desirable to study though costly to tackle with recorded and annotated data. To this end, a simulation protocol was needed, which would be based on a morphological model of environmental acoustic scenes. As discussed in details in Section \ref{sec:discussion}, we acknowledge that the use of simulated data shall not be considered as sufficient for the final evaluation of engineering systems. That being said, the above described potential benefits are still sufficient to justify pursuing that avenue of research. 

The aim of the morphological model proposed in this paper is to generate acoustic scenes as a ``\,skeleton of events on a bed of texture\,''
\cite{nelken_ear_2013}. As the final use of the simulated scenes are to be analyzed by event recognizers trained on recorded data, one shall minimize both the discrepancy between the simulated scenes and recorded ones and its potential impact. Thus, we do not consider approaches based on actual synthesis of sounds. It thus departs significantly from models used in research fields such as wave field synthesis \cite{wavefield}, binaural or spatial scene synthesis \cite{1315647}, acoustic event synthesis \cite{verron20103} and texture synthesis \cite{schwarz2011state, mcdermott_sound_2011,turner2008modeling}.

The proposed model is based on several sequences of sound events issued from the same source, where each sound event is drawn from a collection of carefully chosen sound samples. The morphological aspects of the scene, \textit{i.e.} which sound sample is played at what time and which level, are then modeled in an abstract manner, allowing us to control high level properties of the scene. The contribution of this paper is threefold: 1) propose a computational model for the generation of simulated data sets, 2) motivate important morphological aspects of the model based on perceptual considerations, and 3) consider this simulation paradigm to gain knowledge about the behavior of several event detection systems developed by different research teams  worldwide initially submitted to the DCASE challenge \cite{Giannoulis:2013a}.

To this end, Section~\ref{sec:soundcollection} motivates some design choices, and details the structure of the so called ``\,sound collections\,'', that is, the input data of the simulation process. Section~\ref{sec:model} presents the proposed model of acoustic scenes which underlines the simulation process. Sections~\ref{sec:corpussimulation} and \ref{sec:experiments} present the evaluation framework for event detection systems using simulated acoustic scenes. Then, the use of simulated data to evaluate detection algorithms is discussed in Section~\ref{sec:discussion}.

\section{The notion of sound collection}
\label{sec:soundcollection}
\subsection{Auditory scene as a sum of sound sources}
\label{sec:asa}
%Although major progress have been made in environmental sounds synthesis \cite{schwarz2011state}\cite{verron20103}, current state of knowledge does not allow to adopt a synthesis approach which would consists in synthesizing all the sounds of an entire sound environment.  To tackle this issue, 

As a simulation process couldn't practically deal  with each acoustic event that may occur in an acoustic scene separately, the proposed model adopts a ``\,source-driven\,'' approach by considering an acoustic scene as a sum of sound sources. This approach is consistent with the way humans perceive their sonic environment. Studies addressing the Auditory Scenes Analysis (ASA)\cite{bregman1994auditory} problem, and more specifically the sound segregation process\cite{snyder2007toward}\cite{ciocca2007auditory}\cite{carlyon2004brain}\cite{ballas1987interpreting}, show that humans make sense from their sonic world by isolating information related to individual sound sources. Considering a bottom-up approach, the segregation process relies on generic rules involving Gestalt-like principles \cite{ballas1987interpreting} to group sounds with similar acoustic indicators (common onset, spectral regularity and harmonicity), as well as similar perceptual attributes (timbre, loudness, perceived location and pitch) into perceptual entities called ``\,auditory streams\,''. Recently, several neurophysiological studies have shown evidence of the existence of auditory streams\cite{nelken2008neurons}.

Besides ASA studies which mostly consider pure tones or simple complex sounds \cite{ciocca2007auditory}, more recent studies adopting a psycho-linguistic approach to describe recorded sounds, have also demonstrated the existence of top-down source-driven grouping processes involved in sound perception. Investigating the qualitative evaluation of urban acoustic scenes using categorization tasks and linguistic analysis, studies of Dubois and colleagues \cite{dubois2006cognitive}\cite{raimbault_urban_2005} have shown that listeners categorize sound environments on the basis of semantic features, that is the meaning attributed to the recalled sound sources. 

Considering both the ASA and the psycho-linguistic approach, it seems intuitive for the simulation process to consider separately the sound activity of each sound source of the scene. In practical terms, to materialize these sound activities, each sound source has to be related to a collection of sound recordings. But this approach introduces fundamental questions about the very nature of such a collection. It first questions the existence of a standardized taxonomy of sounds. Such taxonomy must be a hierarchical classification system putting together sounds according to their shared characteristics. Each group must be labeled in a way that a specific name may describe its corresponding class, an instance of it, but also at which level of the classification it fits. Unfortunately, if such systems exist for plants, animals or colors, it is not the case for sounds\cite{niessen_categories_2010}. Main reasons are:
\begin{itemize}
\item Sound description and identification are highly subjective. In other words, a same sound may be described quite differently according to the subject. This is due to the relative lack of basic lexicalized terms to describe acoustic phenomena \cite{guastavino_ideal_2006}
\item Sound description and identification are highly context dependent, that is, sound source identification depends on the nature of the other co-occurring sound sources \cite{gygi2011incongruency, ballas1987interpreting, niessen2008disambiguating}
\end{itemize}

Even if there is no systematic way to build a sound collection, one may take into account some perceptual considerations to guarantee a certain level of ecological validity. Those considerations are addressed in the next sections.

\subsection{Action and Sources}
\label{sec:as}

Event detection tasks evaluate if an algorithm is able to detect a specific set of sound classes. Ideally, to prevent from low generalization capability, the training set of a given class shall be consistent, that is, class exemplars should be representative of the diversity of the sounds suggested by the class label that may occur in the real world. In our case, the class exemplars are the recordings of a sound collection. 

%As there is no systematic classification system for sounds, there is no immediate way to ensure that a sound collection is consistent. 

Some perceptual considerations can be taken into account to guide the collection building process. First, one may look at the way humans classify~/ categorize sounds. As explained by \cite{houix_lexical_2012}, ``\,Categorization is a cognitive process that unites different entities of an equivalent status\,''. Among other categorization strategies, several studies show that humans categorize sounds according to 1) the type of source (agent, object, functions) and~/ or 2) the action~/ movement causing the sounds \cite{guyot1997, gygi2007similarity, marcell2000confrontation, vanderveer1997, ballas1987interpreting, dubois2006cognitive}.

Human categorization occurs at several levels. Rosch \cite{rosch1978cognition} proposed three levels of categorization for real-world objects namely superordinate, basic, and subordinate. The higher the level, the higher is the abstraction degree of the categories. Considering sound perception, Guyot et al. \cite{guyot1997} proposed a framework where listeners identified sound categories of abstract concepts at a supeordinate level (noise generated be a mechanical excitation), action at the basic level (grating, scratching, rubbing) and 
source at the subordinate level (Dishes, Pen sharpening, Door). Although Houix et al. \cite{houix_lexical_2012} found some differences by showing that sounds seem ``\,to be categorized as sound sources first and only second as actions\,'', it appears that source and action are adequate verbal descriptors for category.

One way to make a sound collection consistent is to consider low-level categories as the intra-category diversity decreases with the level. Considering that, one may label a sound collection using a couple ``\,source-action\,'' (\textit{passing-car}), or at least one of the two, in order to minimize the expected diversity of its recordings. Any name referring to higher category levels may lead to sound collections comprising a too large variety of objects. Such definition of collection then raise two issues:

\begin{itemize}
\item building such collections would suppose the availability of a large number of recorded sounds  to be representative of the diversity suggested by the collection label;
\item adopting a data-centered approach, such collections may lead to a misinterpretation of the results of a detection task for someone who did not build them, as the nature of the entities suggested by the collection labels are ambiguous (ex: a sound collection of \textit{traffic sounds} vs. a sound collection of \textit{passing-car}).
\end{itemize} 

Considering the source-action couple is not sufficient. The experimenters must also choose generic labels for the couple. To do so, one may refer to the work of Gaver \cite{gaver1993world} who proposed a phenomenological taxonomy of everyday sounds, the work of Niessen et al. \cite{niessen_categories_2010} who assessed the consensus of categories mentioned in 166 papers of different research domain using linguistic analysis, and recently, the work of Salamon et al. \cite{salamon14} who built a taxonomy of urban sounds based on the work of Brown et al. \cite{brown_towards_2011}. 

This section shows evidence that labeling a class using the source-action nomenclature helps us to reduce the expected intra-class diversity. However, it does not address the issue of inter-class diversity. Indeed, a naive source-driven approach supposes to record in a source-wise way all the sound activities that may occur in an environment. Considering dense environments such as cities or forest, this may raise important practical issues. To circumvent this problem, one may assume that all the sources do not carry the same potential information, and are not required to be recorded separately. 

\subsection{Texture vs. Event}
\label{sec:Informativenes}

The human brain may easily distinguish between a voice sound and a background of other competing sounds \cite{carlyon2004brain}. Considering the example of an urban acoustic scene, global traffic hubbub sounds are typically uninformative, compared with closer human sounds \cite{southworth1969sonic}. 

Maffiolo \cite{maffiolo_caracterisation_1999} showed the existence of two distinct cognitive processes depending on the listener's ability to identify separate sound events. By asking subjects to categorize recordings of urban environments and using linguistic analysis of the verbal descriptions of the categories, she  found two cognitive categories of sound environments called respectively ``\,event sequences\,'' and ``\,amorphous sequences\,''. Event sequences (sound environments in which distinct events or sequences of events can be identified) are processed analytically, that is, based on the meaning of the identified sound sources, whereas amorphous sequences (sound environments in which no event can be isolated) are processed holistically using global acoustical indicators (intensity, spectral content). The distinction observed by Maffiolo was validated by Guastavino \cite{guastavino_ideal_2006}. Using semantic analysis of verbal descriptions of specific sounds populating the urban environment, Guastavino showed that verbal descriptions of low pitched sounds may be divided into two categories called ``\,source events\,'' (sound events which can be attributed to a sound source), and ``\,background noise\,'' (where no identifiable event can be isolated). 

What comes out from these studies is that sound perception highly depends on semantic features (source identification), but also on the informativeness of the isolated source. Sound sources that carry  information of interest are processed separately, whereas the other are processed together in a single stream. %Similar findings seem to be made by auditory neuroscience studies \cite{nelken2008neurons}.

Based on this notion of informativeness, another common distinction is made between two perceptual objects called ``\,sound events\,'' and ``\,sound textures\,''. Based on previous studies on vision, McDermott and Simoncelli \cite{mcdermott_sound_2011, mcdermott2013summary} showed that the perception of sound textures may derive from simple statistics of early auditory representations. These summary statistics would be sufficient to recognize sounds having some temporal homogeneity.

% The authors argue that textures do not require a detailed perceptual analysis in contrast with event perception which leads to semantic categorization involving ``\,a conscious access to a fairly small set of perceptual dimensions\,''\cite{nelken_ear_2013}. 
%
%This may imply that the auditory system would be able to detect whether a sound is a texture or an event before proceeding to the adapted processing. The following sentence of Nelken and Cheveign\'e \cite{nelken_ear_2013} illustrates the last statement:  \\
%
%\begin{quote}
%\textit{... the auditory system may maintain a multi-stream representation involving events and statistics of multiple sources: a skeleton of events on a bed of texture.} \\
%\end{quote} 

That said, there are few formal definitions concerning the texture object \cite{schwarz2011state}. The most notable attempt has been made by Saint-Arnaud \cite{saint1995classification} and Saint-Arnaud and Popat \cite{saint1995analysis}. From their experiment, they derived the following properties (quoted from \cite{saint1995classification}): \\

\begin{itemize}
\item \textit{Sound textures are formed of basic sound elements, or atoms;}
\item \textit{atoms occur according to a higher-level pattern, that can be periodic or random, or both;}
\item \textit{the high-level characteristics must remain the same over long time periods (which implies that there can be no complex message);}
\item \textit{the high-level pattern must be completely exposed within a few seconds (``\,attention span\,'');}
\item \textit{high level randomness is also acceptable, as long as there are enough occurrences within the attention span to make a good example of the random properties.} \\
\end{itemize}

Considering these properties, a texture may be understood as a composite object with two hierarchical levels, the top level being the high level pattern, and the leaf level being the atom. The nature of an atom remains adaptable as the latter may be considered at several time scales. Thus and to some extent, texture may be considered as a concatenation of recordings, each of them being a sequence of atoms. In this case, these recordings must comprise at least the high level pattern of the texture, that is, if we consider a texture of `gallop', recordings of atom sequences  must be at least composed of the first three sounds of hoofs.

To summarize the previous statements, it appears that all sounds are not processed as a sum of distinct events:\\ 

\begin{itemize}
\item amorphous sequences that convey low semantic information are processed holistically;
\item sound textures with stable acoustic properties over long period are processed using summary statistics of these acoustics properties. 
\end{itemize} 

To circumvent the issue of recording a representative number of sound collections to simulate an acoustic scene, one can take into account those considerations, and use recordings of mixed sound sources, provided that they are amorphous sequences or textures. We believe that there exist some links between the notions of amorphous sequences and textures. Both trigger holistic processing based on global acoustical properties for amorphous sequences \cite{maffiolo_caracterisation_1999, dubois2006cognitive} and summary statistics for textures \cite{mcdermott2013summary}, and both convey a low potential information content 
% (see Figure~\ref{fig:sainArnaud}, adapted from 
\cite{saint1995classification}). Yet, amorphous sequences are described as ``\,background sounds\,'' with no identifiable events, whereas  the texture definition comprises sequences of events such as ``\,gallop\,'' that do not meet this last criterion. Considering that, one can consider an amorphous sequence to be a texture, as the physical characteristics of an amorphous sequence remain stable over time, but the reverse is not systematic. 

%\begin{figure}[t]
%\begin{center}
%\includegraphics[scale=0.2]{saintArnaud.png}
%  \caption{\label{fig:sainArnaud} Potential information content of a sound texture vs. time (from \cite{saint1995classification})} 
%  \end{center}
%\end{figure}

\subsection{Definition of a sound collection}
\label{sec:collection}
From the considerations discussed above, we derive two types of sound collections to be used as basic elements by the simulation process: the ``\,event collections\,'' and the ``\,texture collections\,''. For both collections, a stream is modeled as being a temporal sequence of sound recordings coming for the same sound collection. For the texture collection, each recording is an atom sequence, or more precisely, a sequence of sound events which follow a periodic or a stochastic pattern. The nature of the sequence to be recorded depends on the type of texture considered. For a texture with a periodic pattern such as gallop, recordings are event sequences  comprising at least the first three sounds of hooves. And for a texture with a stochastic pattern such as ``\,rain\,'', the recordings are simply samples of rain sounds. This method offers a certain flexibility, as it makes it possible to quickly generate various versions of a same texture with few recorded samples, by varying the apparition order of the 
sequence. 
Obviously for a texture to be realistic, sequences have to come from the same recording session. Moreover, as the human brain is very sensitive to repetition of identical sounds, even when they are individual chunks of white noise \cite{agus_rapid_2010}, a sequence shall not be concatenated with itself. 

To summarize, the proposed source-driven model uses collections as basic element for the simulation process:

\begin{itemize} 
\item Each collection is a group a sound recordings.
\item Insofar as possible, the label of the collection should be of the form ``\,source + action\,'' and labels of source and action must be generic.
\item There are two types of collections called respectively the event collections, and the texture collections.
\item Sound recordings of a same event collection come from the same sound source.
\item Sound recordings of a same texture collection are atomic sequences emitted by one or a mixture of sound sources.
\item Sound recordings of a texture collection must at least comprise the high-level pattern of the texture (\textit{e.g.} three sounds of hooves for the gallop texture).
\item A texture built from the concatenation of recordings must convey a low semantic information and~/ or have stable acoustic properties over time.
\end{itemize}

%To conclude, we believe that the distinction made between events and textures in the literature is not definite, as some sounds may be considered as textures or sound events depending on the studies. Also, most engineering studies dealing with automatic sound event recognition do not use this distinction. They rather distinguish between the foreground events and the background noise, which is closer to the perceptual distinction made between 'event sequences' and 'amorphous sequences'. 

%Thus, for the purpose of this study, \gl{continue...}

\section{Proposed Model} \label{sec:model}
\subsection{Design choices}

\begin{figure}[t]
\centering
\def\svgwidth{\columnwidth}
%% Creator: Inkscape inkscape 0.48.4, www.inkscape.org
%% Accompanies image file 'controlParameters2.pdf' (pdf, eps, ps)
\begingroup%
  \makeatletter%
  \providecommand\color[2][]{%
    \errmessage{(Inkscape) Color is used for the text in Inkscape, but the package 'color.sty' is not loaded}%
    \renewcommand\color[2][]{}%
  }%
  \providecommand\transparent[1]{%
    \errmessage{(Inkscape) Transparency is used (non-zero) for the text in Inkscape, but the package 'transparent.sty' is not loaded}%
    \renewcommand\transparent[1]{}%
  }%
  \providecommand\rotatebox[2]{#2}%
  \ifx\svgwidth\undefined%
    \setlength{\unitlength}{832.06474609bp}%
    \ifx\svgscale\undefined%
      \relax%
    \else%
      \setlength{\unitlength}{\unitlength * \real{\svgscale}}%
    \fi%
  \else%
    \setlength{\unitlength}{\svgwidth}%
  \fi%
  \global\let\svgwidth\undefined%
  \global\let\svgscale\undefined%
  \makeatother%
  \begin{picture}(1,0.37539404)%
    \put(0,0){\includegraphics[width=\unitlength]{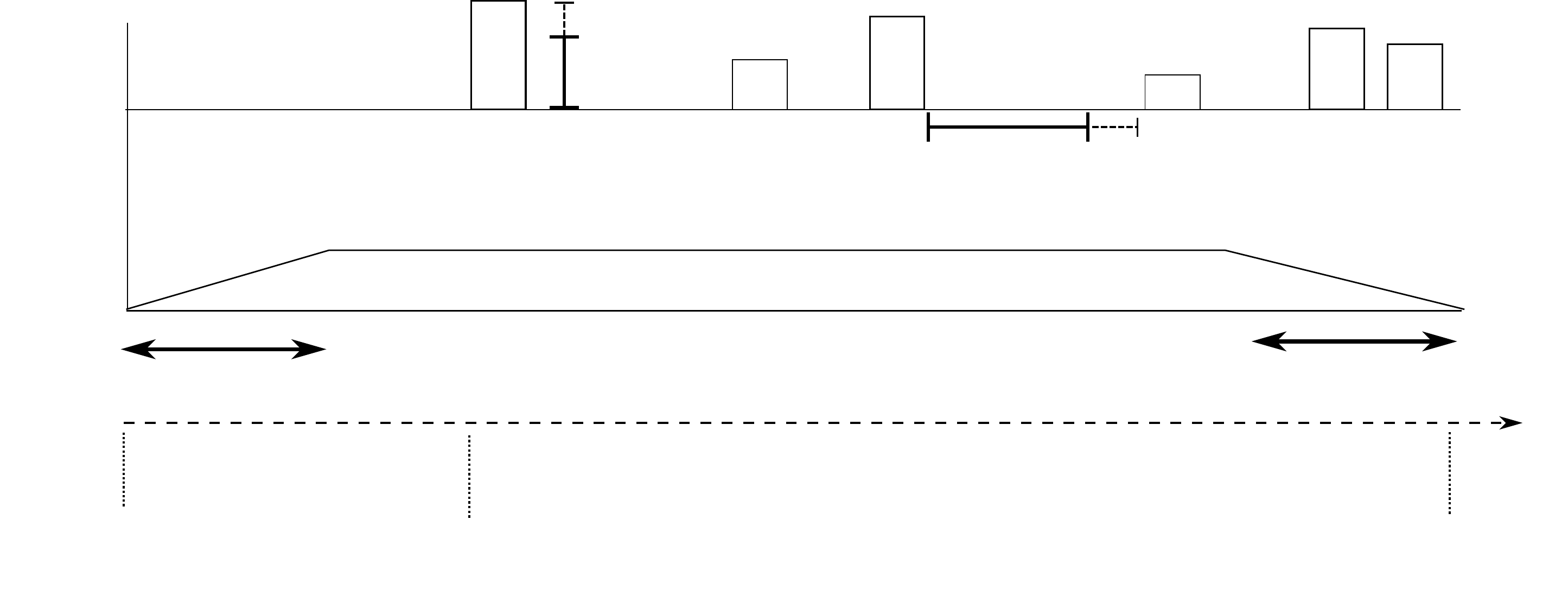}}%
    \put(0.00575811,0.32512232){\color[rgb]{0,0,0}\makebox(0,0)[lb]{\smash{\tiny event }}}%
    \put(0,0.18795398){\color[rgb]{0,0,0}\makebox(0,0)[lb]{\smash{\tiny texture }}}%
    \put(0.45764387,0.07194046){\color[rgb]{0,0,0}\makebox(0,0)[lb]{\smash{\footnotesize  \textit{time}}}}%
    \put(0.07563472,0.12026615){\color[rgb]{0,0,0}\makebox(0,0)[lb]{\smash{\tiny fade In/Out}}}%
    \put(0.08172171,0.02294988){\color[rgb]{0,0,0}\makebox(0,0)[b]{\smash{\tiny Texture}}}%
    \put(0.30597838,0.02294988){\color[rgb]{0,0,0}\makebox(0,0)[b]{\smash{\tiny Event
}}}%
    \put(0.90093639,0.0226837){\color[rgb]{0,0,0}\makebox(0,0)[b]{\smash{\tiny Event and texture}}}%
    \put(0.80635234,0.12026615){\color[rgb]{0,0,0}\makebox(0,0)[lb]{\smash{\tiny fade In/Out}}}%
    \put(0.40981616,0.33324077){\color[rgb]{0,0,0}\makebox(0,0)[lb]{\smash{\tiny EBR}}}%
    \put(0.61450068,0.22251218){\color[rgb]{0,0,0}\makebox(0,0)[lb]{\smash{\tiny time intervals}}}%
    \put(0.36263174,0.31999939){\color[rgb]{0,0,0}\makebox(0,0)[lb]{\smash{\scriptsize $\mu^a_{i}$}}}%
    \put(0.36533509,0.35639){\color[rgb]{0,0,0}\makebox(0,0)[lb]{\smash{\scriptsize $\sigma^a_{i}$}}}%
    \put(0.62814314,0.25892034){\color[rgb]{0,0,0}\makebox(0,0)[lb]{\smash{\tiny $\mu^t_{i}$}}}%
    \put(0.70334704,0.25836979){\color[rgb]{0,0,0}\makebox(0,0)[lb]{\smash{\tiny $\sigma^t_{i}$}}}%
    \put(0.08285587,0.00359984){\color[rgb]{0,0,0}\makebox(0,0)[b]{\smash{\tiny start time}}}%
    \put(0.30283388,0.00359984){\color[rgb]{0,0,0}\makebox(0,0)[b]{\smash{\tiny start time}}}%
    \put(0.89944959,0.00359984){\color[rgb]{0,0,0}\makebox(0,0)[b]{\smash{\tiny end times}}}%
  \end{picture}%
\endgroup%

\caption{\label{fig:controlParameters} Two semantic sound tracks (event and texture) and their controlling parameters.}
\end{figure}
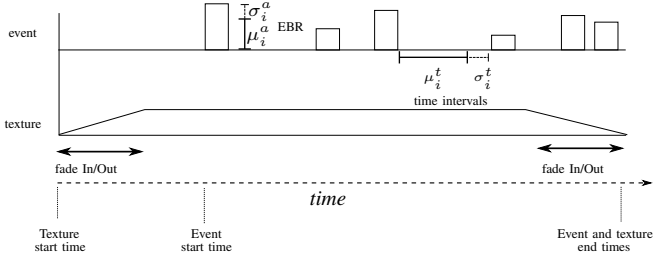

Building on the above discussed matters, the proposed simulation process considers an acoustic scene as a sum of sound sources. Each sound source activity is symbolized as a semantic sound track, which is a sequence of acoustic samples all emitted by the considered sound source (see Figure \ref{fig:controlParameters}). To generate each semantic sound track, the model takes into account a set of four parameters being respectively:

\begin{enumerate}
\item the mean / variance of the  Event to Background power Ratio (EBR) between acoustic samples
\item the mean / variance time interval between consecutive onsets of acoustic samples
\item the mean / variance duration between acoustic samples
\item the start / end times of the track
\end{enumerate}

As motivated in Section \ref{sec:Informativenes}, the model distinguishes between sound events and texture. A track of events is made of discrete sound samples, whereas a track of texture consists of one continuous sound, or a concatenation of samples (see Figure \ref{fig:controlParameters}). Thus, for texture track, the mean/variance time interval between samples as well as the variance EBR are set to $0$. %Evidence that this distinction makes sense from a perceptual view is given in the section \ref{sec:Informativenes}
 
Each semantic track, texture or event, is related to a specific sound collection. As discussed in Section \ref{sec:soundcollection} a sound collection may be seen as a group of similar recordings, each of which comprising sound signals that are emitted by the same sound source. For the purpose of this study, the notion of sound collection greatly overlaps the notion of sound class, as this term is understood when tackling automatic detection tasks.

\begin{figure}[t]
\begin{center}
\includegraphics[scale=0.3]{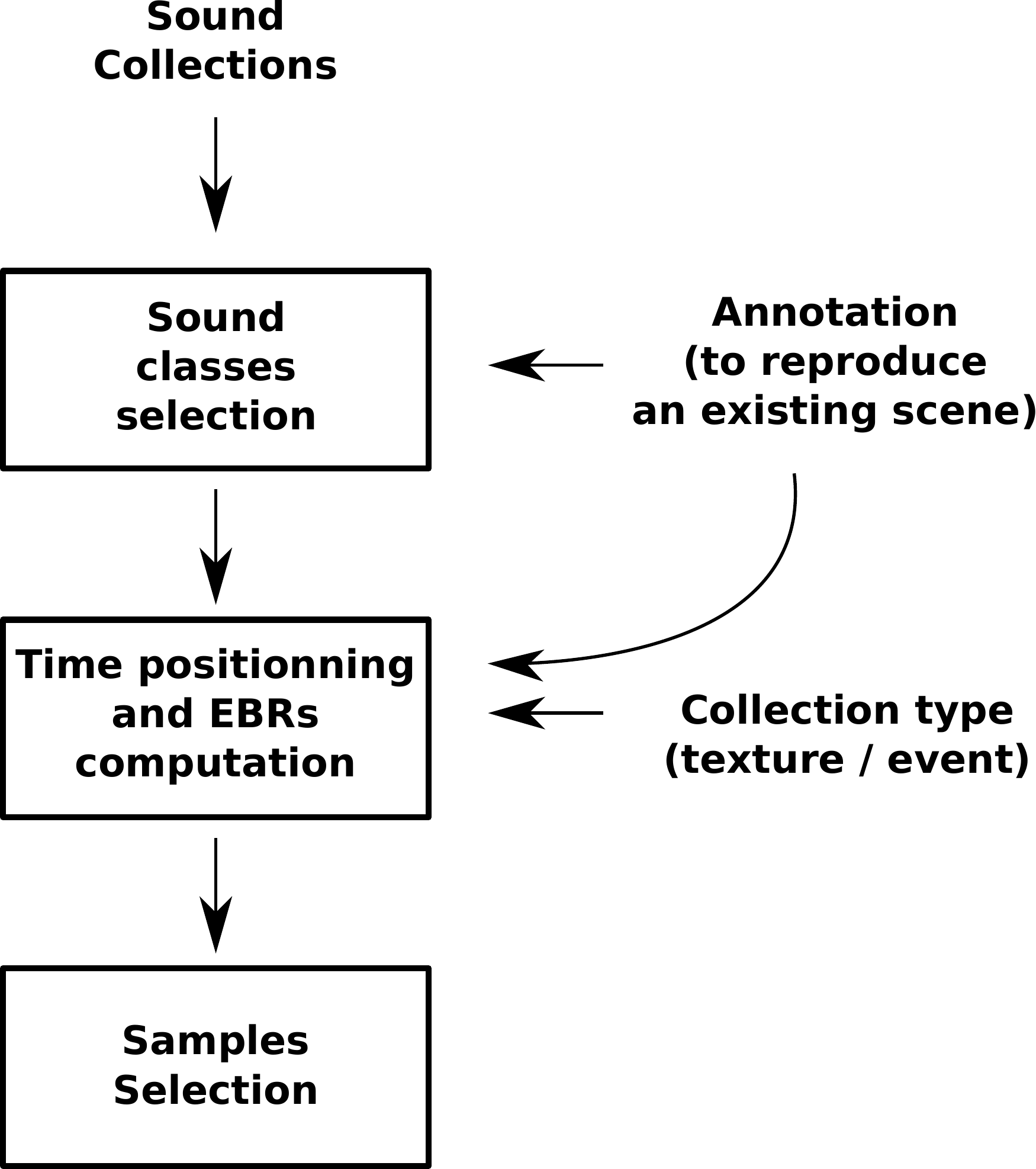}
  \caption{\label{fig:sequencingModel} Schematic of the simulation process.} 
  \end{center}
\end{figure}

The resulting simulation model is depicted on Figure \ref{fig:sequencingModel}. First, the experimenter selects a number of sound sources or class to be used, each of which being related to a specific sound collection. Second, the experimenter sets the simulation parameters depending on the nature of the track (event or texture). Those parameters can also be estimated from pre-existing annotated recorded sound scenes. According to those parameters, the simulation process computes the number of samples used in each semantic track. Lastly, samples are randomly drawn from the corresponding sound collection using a discrete uniform distribution. % The acoustical properties of the recordings, as well as the simulation process depend on the nature  (event or texture) of the sound collection

\subsection{Formal Definition}
\label{Proposed model}

The proposed model is source-driven as it uses  as basic elements semantic sound tracks, gathering sounds coming from the same collection of either sound events or sound textures (see Figure~\ref{fig:sequencingModel}). The nature of the recordings depends on the type of collection (event of texture) which is considered. After selecting the classes to be used, the putative recorded samples are sequenced  to generate the sound environment. The sequencing process depends on the type of collection. Ideally, the sound collection design has to fulfill some perceptual constraints for the simulation to be ecologically valid, \textit{e.g.} to produce realistic scenes, as described in Section \ref{sec:soundcollection}.

Considering that $s(n)$ is a given acoustic scene composed of $C$ sound classes $c_i$, the proposed model is such that:

\begin{equation}
s(n)=\sum_{i=1}^{C}t_{i}(n)
\end{equation}

Where each $n$ is the time index, $t_i$ is a semantic sound track. For the sake of simplicity, we only detail here the model of an event track, then explain the adaptation of the model to texture tracks.

$t_i$ is defined as a sequence of $n_i$ sound events $e_i^k(n)$ randomly chosen among the $|c_i|$ samples in class $c_i$: for each $k$ in $[1..n_i]$, $e_i^k = c_{i, \mathcal{U}(1, |c_i|)}$, where $\mathcal{U} (x, y)$ represents an uniformly distributed integer random value between $x$ and $y$ included. Each event is scaled by an amplitude factor sampled from a real normal distribution with average $\mu^a_{i}$ and variance $\sigma^a_{i}$. The interval separating the onset times of consecutive samples for track $i$ is, similarly, randomly chosen following a normal distribution with average $\mu^t_{i}$ and variance $\sigma^t_{i}$. Formally, each sequence $t_{i}$ is thus expressed as:

\begin{equation}
\label{eq1}
t_{i}(n)= \sum_{j=1}^{n_i} \mathcal{N}(\mu^a_{i},\sigma^a_{i})c_{i, \mathcal{U} (1, |c_i|)}(n-n^j_i)
\end{equation}
\begin{equation}
\label{eq2}
n_i^j=n_i^{j-1} + \mathcal{N}({\mu^t_{i},\sigma^t_{i}})
\end{equation}

where $n_i^0$ is set to $0$ by convention. The signal of an event is defined in such a way that $e(n)=0$ if $n<0$ or beyond the signal's duration. 

In the case of a texture track, two implementation differences must be observed to maintain a perceptually acceptable output: first, signal amplitude is only drawn at random once, and that value is applied to all samples; second, sample start times are not randomized but chosen so that the texture recordings chosen from class $c_i$ will be played back-to-back with sufficient overlap to create an equal-power cross-fade between them, thus generating a continuous, seamless track.

While implementing this model for the generation of acoustical sound scenes, additional treatments and constraints are applied in order to improve perceptual quality. Namely, the fading of the onset and offsets of events and the whole track, and the fact that the same sound sample cannot be sequenced consecutively.

\section{Corpus simulation}
\label{sec:corpussimulation}

%\ml{Petite intro sur l'etape de validation du model}

This section describes the different corpora of simulated acoustic scenes considered in the experiments described in Section \ref{sec:experiments}. All the scenes are simulated using the DCASE challenge test set annotations for the `Office Live' (OL) task \cite{giannoulis2013detection}.  We run the same automatic event detection algorithms used for the DCASE challenge, and compare the results obtained with the simulated scenes to those obtained with the real scenes of the DCASE test set. 

%The goal is twofold proved that an event based model is relevant to simulate credible sound environments.

The root corpus is the test set considered in the DCASE challenge. It is called ``\,test-QMUL\,''. This corpus is composed of 11 recordings of office live scenes  roughly one minute long. Scenes have been recorded in 5 different acoustic environments. The audio events have been divided into 16 sound event classes to be annotated: door knock, door slam, speech, human laughter, clearing throat, coughing, drawer, printer, keyboard click, mouse click, object (specifically pen, pencil or marker) put on table surfaces, switch, keys (put on table), phone ringing, short alert (beep) sound and page turning. Two different annotations coming from two distinct individuals have been used to measure the algorithm performances, thus leaving us with 22 scene-annotator couples. There is no time overlap between events.

\begin{figure}[t]
\begin{center}
\includegraphics[scale=.44]{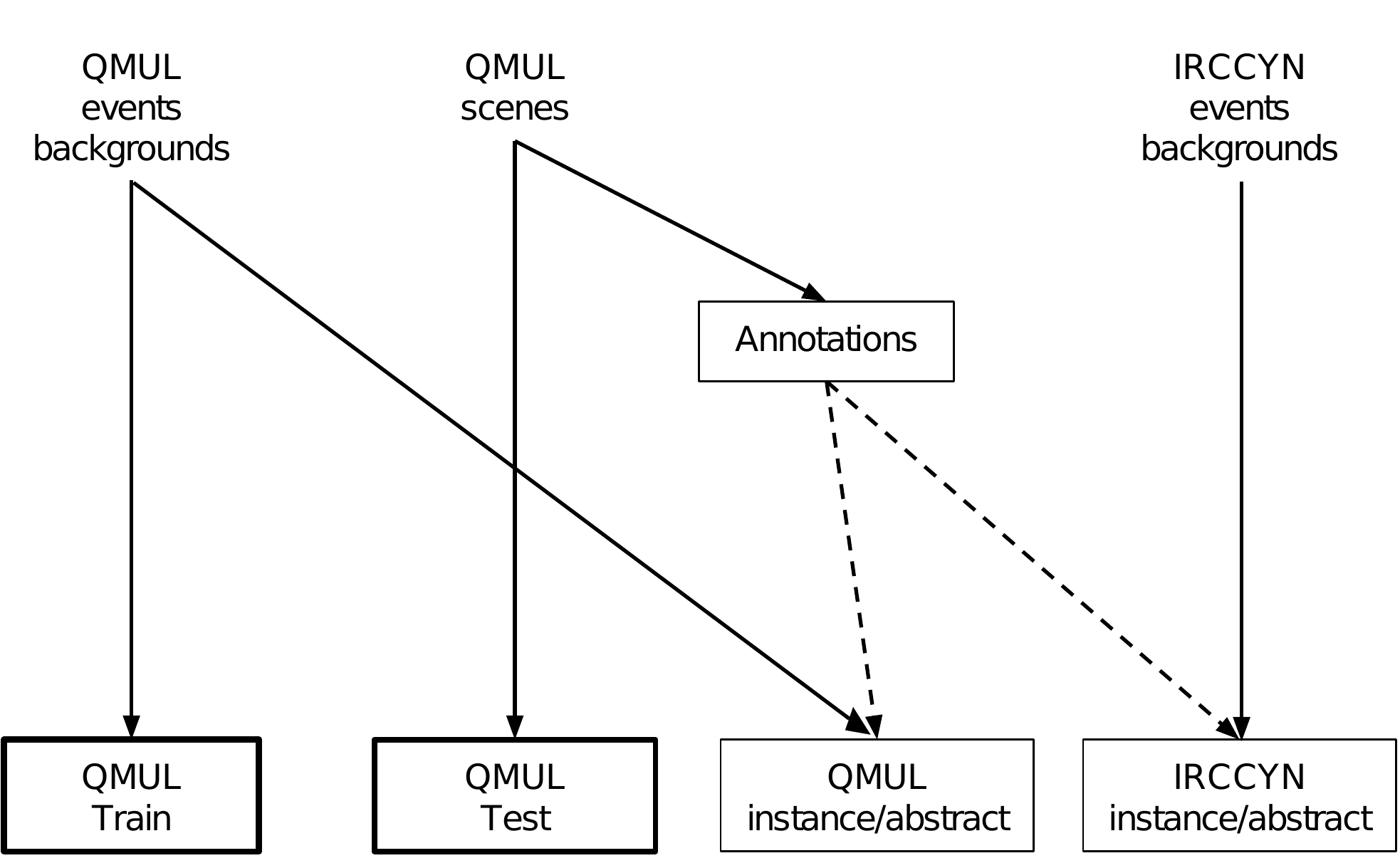}
  \caption{\label{fig:databases} Generation process of the corpora considered in this evaluation. As part of the DCASE challenge, systems were trained on QMUL Train and tested on QMUL Test during the DCASE challenge.} 
  \end{center}
\end{figure}

Four corpora of simulated scenes are generated as depicted in Figure \ref{fig:databases}. They are respectively called ``\,instance-QMUL\,'', ``\,abstract-QMUL\,'', ``\,instance-IRCCYN\,'' and ``\,abstract-IRCCYN\,''. The labels  ``\,IRCCYN\,'' and ``\,QMUL\,''  refer to the  two different datasets of event recordings used to generate the corpora which have been recorded in different offices, the ones of Queen Mary University of London (QMUL) for the former and the ones of the Insitute of Research on Communications and Cybernetics of Nantes (IRCCYN). The labels  ``\,instance\,'' and ``\,abstract\,'' correspond to two distinct simulation processes. 
%The corpus of  real scenes used in the DCASE challenge is called ``\,test-QMUL\,'', as its have been considered to evaluate the detection algorithms during the DCASE challenge.

To generate the two QMUL corpora, we use recordings of audio events that have been extracted from recordings done during the preparation of the DCASE challenge, but unused during the challenge, see \cite{Giannoulis:2013a} for further information on recording conditions. The extracted samples were therefore recorded in the same conditions than the test-QMUL corpus. Depending on the sound class considered,  3 to 23 events per class are extracted. We also use event-free background recordings (texture) coming from the same acoustic environments than those of the test-QMUL corpus. These background recordings are used to generate the background noise (texture) of the instance-QMUL and abstract-QMUL corpora. % Each event-free background recording was long enough to avoid the need for concatenation. 

The two IRCCYN corpora are generated using new recordings of sound events with respect to the sound classes of the DCASE challenge. All recordings were performed at IRCCYN in a calm environment using the shotgun microphone AT8035 connected to a ZOOM H4n recorder. 20 samples of each class are used to generate the instance-IRCCYN and abstract-IRCCYN corpora, which corresponds to the cardinality of the DCASE challenge train set in terms of event classes \cite{Giannoulis:2013a}.

%The following sections describes the two simulation processes ``\,instance\,'' and ``\,abstract\,''. 

\subsection{``\,Instance\,'' simulation process}
\label{sec:instance}

The instance simulation process simulates acoustic scenes with the same temporal structure and Event to Background Ratios (EBRs) than the annotation of the test-QMUL corpus. The EBR of an event of $N$ sample length is obtained by computing the ratio in decibel between the event $E_{rms}$ and the background $B_{rms}$ root means square measures.\\

\begin{equation}
EBR = 20log_{10}\left( \frac{E_{rms}}{B_{rms}} \right)
\end{equation}\\
with 
$$X_{rms}=\left({\dfrac{1}{N}\sum\limits_{n=1}^{N}x(n)^{2}}\right)^{1/2}$$
 x(n) may be replaced by $e(n)$ and $b(n)$, the sound pressures at sample $n$ of respectively the sound event and the background noise. \\

For each event of each scene-annotator couple of the test-QMUL corpus, the onset-offset times and an approximation of the EBR are considered. As it is not possible to isolate the background under the events, the background level needed to compute the EBR is obtained using a event-free sequence of each real scene. These onsets-offsets  and EBR are then used to generate the simulated scenes. For each simulated scene, at each onset of the corresponding annotator-couple scene, we randomly place an audio event  belonging to the same audio class. To ensure that samples of recorded audio events are not too long comparing to the annotated ones, recordings are cut off to the annotation length if the recording duration is larger than the annotation duration of at least 0.5 seconds. 

Each event has its amplitude scaled to the same EBR than the test-QMUL corpus. Instance simulation process provides us with  simulated scenes with temporal structures and sound levels that are close as possible as those of the real corpus test-QMUL.\\

\subsection{``\,Abstract\,'' simulation process}
\label{sec:abstract}

For the abstract simulation process, the goal is to abstract temporal structures and EBRs of the real scenes. To do so, the model described in \ref{Proposed model} is instantiated using estimations of the $\mu^a_{i}$, $\sigma^a_{i}$, $\mu^t_{i}$ and $\sigma^t_{i}$ parameters  (see eq. \ref{eq1} and \ref{eq2}). Estimation is done for each annotator-scene couple, using both the sound signals and the annotations of the test-QMUL corpus. 
To generate the simulated scenes, EBRs and time intervals between events are respectively obtained from the Normal distributions $\mathcal{N}(\mu^a_{i},\sigma^a_{i})$ and $\mathcal{N}({\mu^t_{i},\sigma^t_{i}})$.

Similarly to the instance simulation process, event recordings are chosen randomly.  For practical considerations, the start and termination times of the class sequence (semantic sound track) are the same as the ones of the test-QMUL corpus. To ensure that the recorded samples are not significantly longer compared to the annotation times, the sample duration of a considered sound class $i$ has its duration $D$ thresholded as follows: $D-\mu_i-\sigma_i>5$, with $\mu_i$ and $\sigma_i$ being respectively the average and standard deviation of the duration of the events belonging to the class $i$ in a given annotation. Setting the  lower bound to 5 seconds allows us to minimize the impact of such operation on short impulsive sounds.

\section{Experiments} \label{sec:experiments}

\subsection{Evaluation Metric}

The performance of event detection systems can be evaluated following several metrics. In order to improve legibility of the following, we shall retain one evaluation metric that is considered to be the most informative for our study.

Among the four metrics considered in the DCASE Challenge\cite{Giannoulis:2013a}, namely the Acoustic Event Error Rate (AEER) \cite{clear}, the Precision, Recall, and F-measure, the F-measure is selected as the most common and interpretable one.

Another variation is that those metrics can be computed over each frame or on event boundaries. In the latter case, the detection of the onset boundary can be considered solely or together with the offset. As annotating and consequently detecting the duration and the offset of events is notoriously difficult, we focus on the detection of the onset as the main objective. Furthermore, in order to achieve more comparable results across datasets and to ensure that repetitive events do not
dominate the accuracy of an algorithm, the metric shall be class normalized. That is:
\begin{equation}
f = \frac{1}{C} \sum_{i=1}^{C} f_i 
\end{equation}
where $f_i$ is the F-measure achieved by the system while detecting event $i$. Thus, by considering the Class-Wise Event onset based F-measure (CWEBF), performance evaluation is  more invariant to event duration and distribution. We thus select this metric that was also collectively agreed upon by DCASE participants through the challenge mailing list.

\subsection{Detection systems}

\begin{figure*}
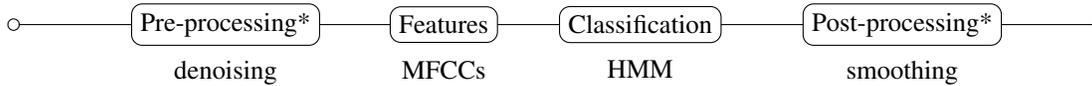

\center
\tikzset{
    mynode/.style={rectangle,rounded corners,draw=black, top color=white, text centered},
} 
\tikz \draw [o->] (0,0) -- (.8\textwidth,0)
node[mynode, pos=0.2] {Pre-processing*} 
node[mynode, pos=0.4]  {Features}
node[mynode, pos=0.58]  {Classification} 
node[mynode, pos=0.82] {Post-processing*} 
node[pos=0.2, below=10pt] {denoising} 
node[pos=0.4, below=10pt] {MFCCs} 
node[pos=0.58, below=10pt] {HMM} 
node[pos=0.82, below=10pt] {smoothing};
\caption{Schematic of event detection systems (nodes with a * are not systematically used). Below, state-of-the-art design choices are given as examples.}
\label{fig:schematic}
\end{figure*}

Together with a Baseline system provided by the organizers, 8 detection systems have been evaluated at the DCASE challenge. Those systems roughly follow the processing chain shown on Figure \ref{fig:schematic} with some variety on the implementation of the different nodes. Features are most commonly Mel-Frequency Cepstral Coefficients (MFCC)s \cite{Davis80a} but other sets fo spectral features are also considered, with or without pre-processing such as denoising. The classifier of choice is the 2 layer Hidden Markov Model (HMM) \cite{18626} where the second layer models the transition between events but other classifiers are also considered such as Random Forests (RF), Support Vector Machines (SVM) or Non-negative Matrix Factorization (NMF).

All those algorithmic differences as well as their specific tuning result in specific behaviors that are interesting to evaluate in different testing conditions, especially those which evaluate their generalization capabilities.

\begin{table}[t]
\caption{Summary of submitted event detection systems.}
\label{tbl:EDall}
\vspace{-1.5ex}
 \begin{center}
 % \resizebox{245pt}{!}{
   \begin{tabular}{|l|c|} \hline 
    \textbf{System}  & \textbf{Method}  \\ \hline
   CPS\hfill  \cite{CPS}  & Segmentation - Likelihood ratio test classification  \\ \hline
    DHV \hfill \cite{DHV} & MFCCs (features) - HMMs (detection) \\ \hline  
    GVV \hfill\cite{GVV} & NMF (detection) - HMMs (postprocessing) \\ \hline
    NVM\hfill \cite{NVM}  & Hierarchical HMMs + Random Forests (classification) \\ \hline        
    NR \hfill \cite{NR2}  & MFCCs (features) - SVMs (classification) \\ \hline  
    SCS \hfill\cite{SCS} & Gabor filterbank (features) - HMMs (classification) \\ \hline     
   VVK \hfill\cite{VVK}   & MFCCs (features) - GMMs (detection)  \\ \hline 
    Baseline \hfill\cite{Giannoulis:2013a} & NMF with learned bases (detection) \\ \hline      
   \end{tabular}
  %}
 \end{center}
\vspace{-3.5ex}
\end{table}

\subsection{Datasets}

Five corpora called respectively test-QMUL (described in Section \ref{sec:corpussimulation}), instance-QMUL, abstract-QMUL, instance-IRCCYN and abstract-IRCCYN are used for the evaluation. This section details the corpus specificities. Figure \ref{fig:databases} illustrates the corpora generation process.

All corpora with the labels ``\,QMUL\,'' or ``\,IRCCYN\,'' are obtained using respectively the QMUL or IRCCYN event datasets (see section \ref{sec:corpussimulation}). Similarly, all corpora with the labels ``\,instance\,'' or ``\,abstract\,'' are obtained using respectively the instance or abstract simulation process (see Sections~\ref{sec:instance} and \ref{sec:abstract} resp.).

The instance-QMUL corpus is composed of 4 sub-corpora called respectively ``\,insQ-EBR\_6\,'', ``\,insQ-EBR\_0\,'', ``\,insQ-EBR\_-6\,'' and ``\,insQ-EBR\_-12\,''. For theinsQ-EBR\_0 sub-corpus,  EBRs of test-QMUL scenes  are preserved. To measure the impact of different EBRs on the algorithm performances, we generate three other sub-corpora (insQ-EBR\_6, insQ-EBR\_-6 and insQ-EBR\_--12) by adding three offsets, one for each sub-corpora, to the EBRs of the test-QMUL scenes. The offsets are of $+6$dB, $-6$dB and $-12$dB.

For all the sub-corpora of the instance-QMUL corpus (insQ-EBR\_6, insQ-EBR\_0, insQ-EBR\_-6 and insQ-EBR\_-12) as well as the other corpora (abstract-QMUL, instance-IRCCYN, abstract-IRCCYN), each scene is simulated 10 times, each time using different recording instances. Each of these corpora~/ sub-corpora is composed of 220 simulated scenes ($22*10$) corresponding to the 22 scene-annotator couples of the test-Q scenes replicated 10 times.

\subsection{Results on QMUL datasets}

Granted with permission of the authors of the submitted systems, we ran the above described systems on the simulated datasets on the same computing servers as the ones used for the challenge with the same computing environment. Also, a rerun of the systems over the QMUL Test set has been done in order to ensure replication of the published results.

Table \ref{tab:qmul} shows the class-wise event based F-measure in percent achieved by the evaluated systems over the QMUL Test set and the simulated sets QMUL Instance and QMUL Abstract. The baseline, CPS, GVV and SCS systems performed equivalently across the 2 datasets. The DHV system performed better, but not by a significant margin. The VVK, NVM, and NR2 systems have their performance decreased, by a significant margin for the latter two. The CPS system submitted to the DCASE Challenge had an implementation issue that prevent it to run correctly at the time of the challenge, giving poor results that are consistently replicated over the simulated datasets. For this reason, the CPS system will not be discussed further in the remaining of the paper. Leaving aside the NR2 and NVM systems, the ranking of the systems are same for the 3 datasets. This result comfort us with the use of such simulation scheme to replicate and extend evaluation results achieved on recorded and annotated datasets.

\begin{table} 
\begin{center} 
%\small 
% \setlength{\tabcolsep}{.16667em} 
\begin{tabular}{lccc} 
dataset & testQ & insQ & absQ \\ 
\hline 
Baseline & \textbf{ 9.0$\pm$4.8} & \textbf{10.5$\pm$3.0$^*$} & \textbf{ 9.9$\pm$3.5} \\ 
CPS & \textbf{0.7$\pm$0.8} & \textbf{0.8$\pm$1.3} & \textbf{0.8$\pm$1.4$^*$} \\ 
DHV & \textbf{30.7$\pm$8.4} & \textbf{34.5$\pm$7.5$^*$} & \textbf{34.0$\pm$7.9} \\ 
GVV & \textbf{13.2$\pm$8.0} & \textbf{15.0$\pm$6.4$^*$} & \textbf{14.6$\pm$6.2} \\ 
NR & \textbf{21.5$\pm$6.5$^*$} &  6.8$\pm$5.7 &  7.4$\pm$5.8 \\ 
NVM & \textbf{28.2$\pm$5.9$^*$} &  9.7$\pm$9.6 & 10.8$\pm$9.9 \\ 
SCS2 & \textbf{41.5$\pm$7.6$^*$} & \textbf{39.3$\pm$8.2} & \textbf{39.4$\pm$8.2} \\ 
VVK & \textbf{24.6$\pm$6.8$^*$} & \textbf{19.7$\pm$8.7} & \textbf{19.2$\pm$9.2} \\ 
\end{tabular} 
\end{center} 
\caption{Results of the evaluated systems on the QMUL datasets. Results in bold are equivalent per row(paired t-test at 0.05 significance level) to the best performance per row (depicted with a $^*$).} 
\label{tab:qmul} 
\end{table}

\begin{table}
\begin{tabular}{cllllll}
   System &   testQ &          insQ &   absQ          \\
 \hline
 Baseline & 3.14    (drawer) &  8.63    (drawer) &  7.40    (drawer) \\
      CPS & 2.66    (knock) &  9.04  (doorslam) &  7.84  (doorslam) \\
      DHV & 8.44    (drawer) &  6.88    (drawer) &  8.01  (keyboard) \\
      GVV & 3.08  (pageturn) &  3.78  (pageturn) &  3.55  (pageturn) \\
      NR & 4.33  (keyboard) & \textbf{25.35}  (doorslam) & \textbf{20.68 } (doorslam) \\
      NVM & 1.26  (laughter) & \textbf{22.48}     (cough) & \textbf{19.22    } (cough) \\
     SCS & 1.18     (alert) &  2.70    (drawer) &  1.72  (doorslam) \\
      VVK & 1.81     (alert) &  8.73  (doorslam) &  8.20  (doorslam) \\
\end{tabular}
\caption{Maximal number of false positive averaged over scenes and corresponding event.}
\label{tab:fp}
\end{table}

We shall now investigate further the reasons explaining the behavior of the NVM and NR2 systems. In test mode, both systems first compute features and then run a classifier on them. Therefore,  features were first checked for inconsistent values. The minimal and maximal values did not change across datasets, and the distribution of the features are indeed different across datasets but not by a large margin. 

Close inspection of the inter-class confusion matrices for each systems revealed that for the two systems the classification node may be responsible for this performance. Indeed, one event is triggered almost all the time which drastically increase the false alarm rate. This explains for a large part the decrease of performance of the NR and NVM systems. This behavior can easily be seen on Table \ref{tab:fp} which displays the maximal number of false positive averaged over scenes of each datasets and their corresponding event; on the simulated datasets, the doorslam event for NR and cough event for NVM are falsely triggered very often.

We conclude that this decrease of performance is most probably not due some potential synthesis inconsistencies produced by the simulation process but more due to an over fit of the classification node. Considering that both systems are the only submissions based on discriminative approaches, SVMs and RFs for the NR2 and NVM respectively, we may conjecture that the training framework of the DCASE challenge is not well suited for such classification schemes.

\begin{figure}[t]
\begin{center}
\includegraphics[width=.5\textwidth]{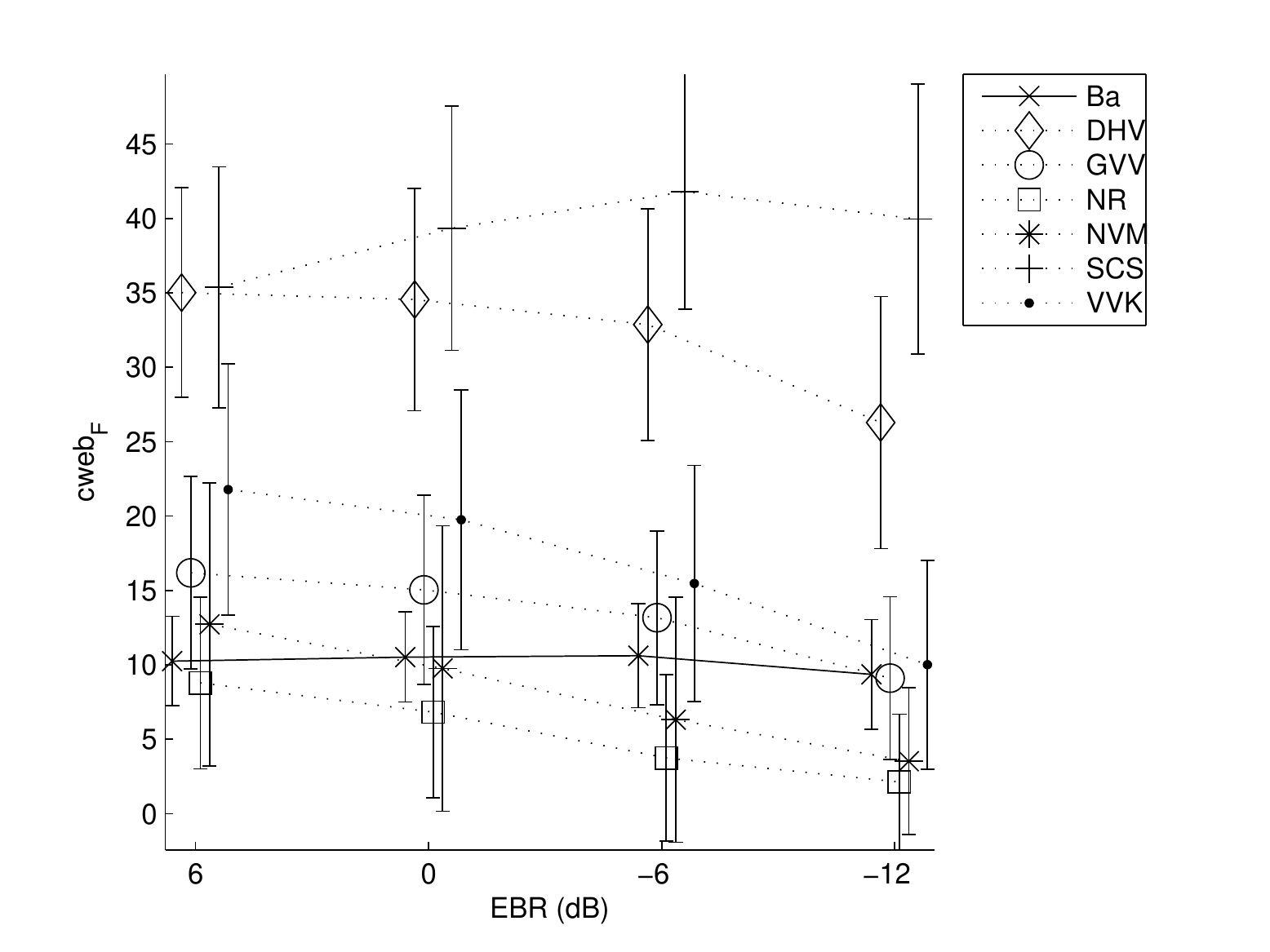}
  \caption{\label{ebr} Class wise event based F-measure in percent achieved by the systems on the QMUL instance datasets with varying EBR.}
  \end{center}
\end{figure}

System performance when the EBR is varied is now studied. Figure \ref{ebr} shows their performance, where 0 dB of EBR roughly corresponds to the EBR level of the QMUL Test set. As expected, most of the systems have their performance decreasing with respect the decrease of the EBR. The ranking is preserved, and the spread between the 3 lowest performing systems greatly reduces with respect the decrease of the EBR. The only system that does not follow this trend is the SCS system, which maintains a stable performance across all EBR ranges. This may be due to an effective signal enhancement which is an important pre processing node of this system \cite{SCS}.

\subsection{Results on IRCCYN datasets}

When tackling a classification task, an important issue is whether the classification system under evaluation is able to generalize to unseen data whose annotation is consistent with the one used for training and tuning.

To evaluate this generalization capability, it is useful to consider results achieved by the systems on the IRCCYN datasets, where the background and events are recorded in a different environment than the one used for recording the training data. 

\begin{table} 
\begin{center} 
%\small 
% \setlength{\tabcolsep}{.16667em} 
\begin{tabular}{lccc} 
dataset & testQ & insI & absI \\ 
\hline 
Baseline & \textbf{9.0$\pm$4.8$^*$} & 5.9$\pm$2.9 & 5.6$\pm$2.9 \\ 
DHV & \textbf{30.7$\pm$8.4$^*$} & 10.0$\pm$5.8 &  9.5$\pm$5.6 \\ 
GVV & \textbf{13.2$\pm$8.0$^*$} &  5.6$\pm$3.7 &  5.5$\pm$3.6 \\
NR & \textbf{21.5$\pm$6.5$^*$} &  4.6$\pm$3.4 &  5.4$\pm$4.5 \\ 
NVM & \textbf{28.2$\pm$5.9$^*$} &  3.1$\pm$3.1 &  3.2$\pm$3.0 \\ 
SCS & \textbf{41.5$\pm$7.6$^*$} & 35.4$\pm$7.2 & 34.0$\pm$6.7 \\ 
VVK & \textbf{24.6$\pm$6.8$^*$} &  6.6$\pm$5.7 &  7.3$\pm$6.3 \\ 
\end{tabular} 
\end{center} 
\caption{Results of the evaluated systems on the IRCCYN datasets. Results in bold are equivalent (t-test per row at 0.05 significance level) to the best performance (depicted with a $^*$).} 
\label{tab:irccyn} 
\end{table}

\begin{figure}[t]

\includegraphics[width=.5\textwidth]{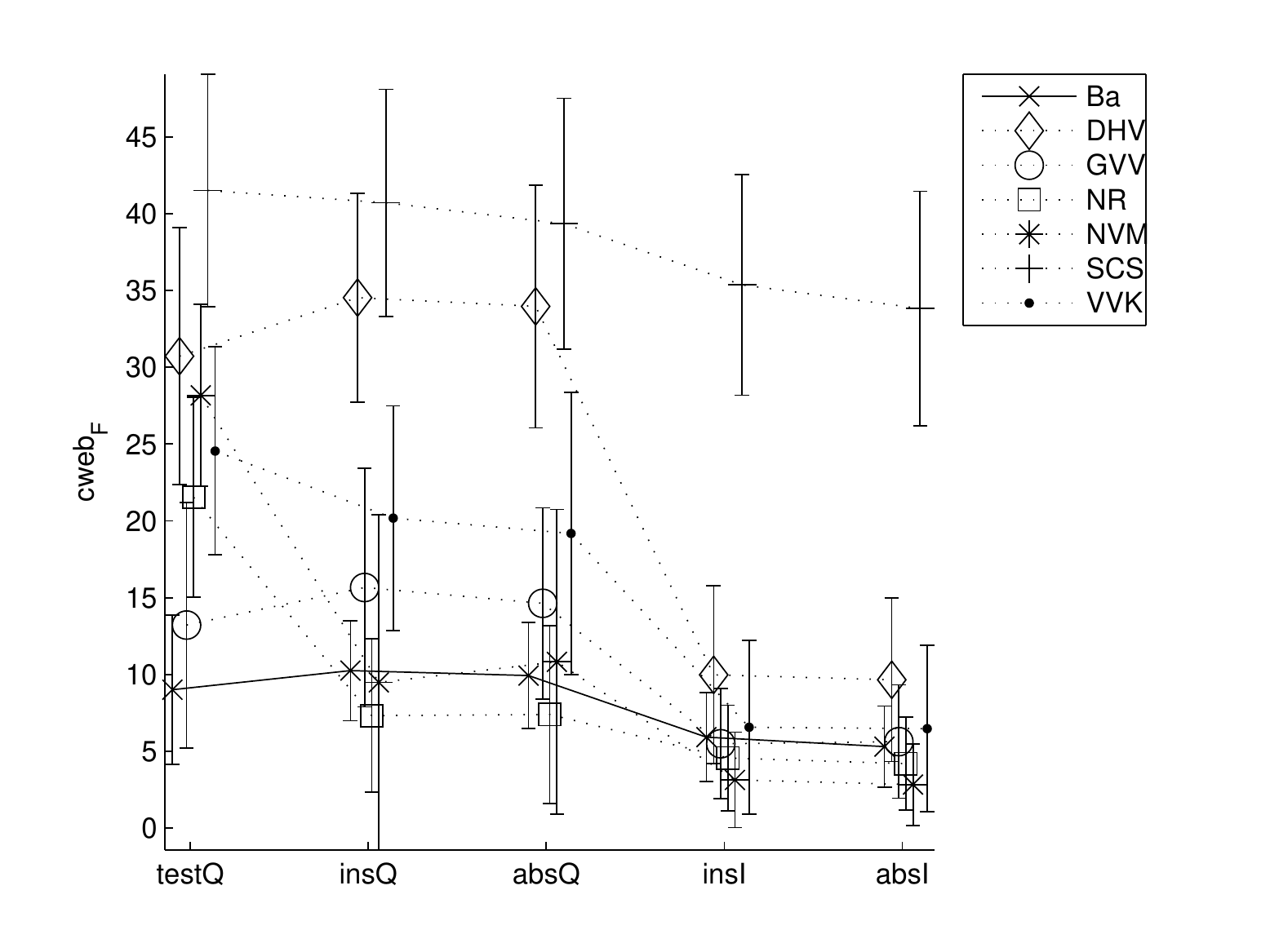}
  \caption{\label{irccyn} Class-wise event based F-measure achived by the different systems on the QMUL and IRCCYN datasets.}
\end{figure}

Whereas the expected behavior with the use of the QMUL instance and abstract datasets was a equivalent performance compared to the ones achieved on test QMUL, the expected behavior with the IRCCYN dataset is a drop in performance. As can be seen on Table \ref{tab:irccyn}, this drop is significant for all the systems. More importantly, all the systems except the SCS one  achieve similar performance when compared to the baseline on the IRCCYN datasets, meaning that for most systems, the performance gain may solely be due to an over adaptation of the system to the training data. Figure~\ref{irccyn} summarizes the results, where the good behavior of the SCS system can be clearly seen.  

\section{Discussion} \label{sec:discussion}

%Whether there is a need to effectively segregate acoustic events is an open question dependant on the task and the algorithmic approach taken by the systems designed to solve such task. Though, recent physiological studies show that such segregation appears in many layers of the auditory systems and even at the lowest ones \cite{Mesgarani2012}, indicating a need for at least considering enhancement of the relevant parts of scene information as an important step to handle acoustic scenes with realistic complexity.

In the light of the results discussed above, we believe that considering carefully designed simulated data is useful for gaining knowledge about the properties and behaviors of the systems under evaluation, thus helping the designer in his algorithmic choices and their evaluation. Important factors influencing the performance such as the noise level, the level of polyphony, the intra-class diversity (acoustical difference between training and testing data) can be evaluated independently, without the burden of experimentally recording data which the desired properties and manually annotating them.

Even though the sole use of synthetic data for validating a computational approach is clearly not sufficient, we believe that the sole use of real data may not be sufficient either, should one wish to gain deep knowledge about the impact of some design  and parametrization issues involved in the implementation of an engineering system.

Indeed, real data which is well annotated is most of the time a scarce resource as the careful design of a large evaluation dataset is a demanding task. Moreover, depending on the task at hand, which may not be always well posed, the annotation can be a critical issue leading to some compromise that will greatly contribute to the difficulty of evaluating the performance of the algorithms.

%We shall stress the difference we make between synthesized data and simulated data. In this case, additive mixture are synthesized data, meaning that the model 

We thus believe that considering simulated data is an in between approach, that together with final validation using real data may be very useful in order to produce more knowledge about the engineering systems under evaluation. We shall stress that such approach is taken in more mature fields, for example robust ASR where challenges are conducted using simulated data such as the CHIME challenges \cite{chime1, chime2}. The simulated acoustic scenes datasets have been generated using a dedicated set of Matlab functions publicly available\footnote{\url{https://bitbucket.org/mlagrange/simscene/downloads}}.

\section{Conclusion}

A morphological model of acoustic scenes has been presented. Following a collection based approach, it generates a set of sound tracks which are sequences of event realizations drawn from specifically tailored sound sample collections. Its potential for generating simulated corpuses of office events scenes is evaluated, by building upon the results obtained thanks to the IEEE AASP DCASE challenge on the detection of events in an office environment. 

We believe that considering those simulated corpora allows us to gain important knowledge about the behavior of the systems under evaluation. As most of the systems under evaluation were built for monophonic inputs (one event occurring at a given time), this paper focused on modifying the acoustical properties of the background or the events. Future research will focus on the influence of the degree of overlap when facing polyphonic scenes, potentially with temporal interactions between events, both for single events (e.g. repetitions for a single event) as well as interactions between event classes.
%
%\gl{one concept that was not discussed in the paper and could be added as future work would be to model temporal interactions for acoustic scenes}

% or


%\appendices

\section*{Acknowledgment}

The authors would like to thank Mark Plumbley for his support. Research project partly funded by ANR-11-JS03-005-01.

\ifCLASSOPTIONcaptionsoff
  \newpage
\fi

\end{document}